\documentclass[10pt,twocolumn]{article}
\usepackage[utf8]{inputenc}
\usepackage{graphicx}
\usepackage{amsmath, amssymb}
\usepackage[numbers]{natbib}
\usepackage[margin=0.75in]{geometry}
\usepackage{hyperref}
\usepackage{microtype}
\usepackage{dblfloatfix}
\usepackage{parskip}
\graphicspath{{./}}
\DeclareMathOperator{\Var}{Var} 

\title{Weight Initialization and Variance Dynamics in Deep Neural Networks and Large Language Models}
\author{Yankun Han \\ University of Florida}
\date{}

\begin{document}
\maketitle

\begin{abstract}
Weight initialization governs signal propagation and gradient flow at the onset of training. This paper presents a theory-grounded and empirically validated analysis across two regimes: compact ReLU MLPs and GPT-2--style transformers. First, a standard-deviation sweep shows that inappropriate scales induce vanishing or exploding gradients, while a broad band around $\sigma\!\in\![10^{-2},10^{-1}]$ yields stable learning. Second, a controlled comparison confirms that Kaiming initialization converges faster and more stably than Xavier under ReLU. Third, I track layerwise weight-variance dynamics in a GPT-2--style model, revealing depth-dependent equilibration toward narrow variance bands. These findings unify classic variance-preserving intuition with modern LLM behavior and suggest practical recipes for robust training.
\end{abstract}

\section{Introduction}
Initialization determines whether deep networks train efficiently or collapse into saturation or instability. Classical prescriptions (LeCun, Glorot/Xavier, He/Kaiming) aim to preserve signal magnitude across layers. However, open questions remain regarding (i) sensitivity to deviations from ideal scales, (ii) rectifier-specific effects under ReLU or GELU, and (iii) whether variance stabilization persists in large transformers.

This paper provides a unified treatment. Experiment E1 quantifies sensitivity to the initial standard deviation (std) via a logarithmic sweep; E2 compares Xavier and Kaiming initialization under ReLU MLPs; and E3 focuses on the statistical characteristics of a GPT-2 model built from scratch during pre-training. It systematically analyzes the evolution of QKV weight variance across all 12 Transformer layers and, using a representative layer as an example, characterizes the changes in its weight-matrix distribution throughout training.

\paragraph{Contributions.}(1) \textbf{Unified variance analysis.} I derive forward and backward variance propagation conditions for common rectifiers (ReLU, GELU), showing when fan-in scaling preserves signal magnitude and how the ReLU factor-of-two arises.
(2) \textbf{Sensitivity to scale.}
Via a logarithmic sweep over $25$ standard-deviation values in ReLU MLPs,
we map the vanishing/exploding regimes and identify a robust training band
$\sigma \in [10^{-2}, 10^{-1}]$, quantified using loss trajectories, accuracy, and gradient-stability indicators.
(3) \textbf{Rectifier-aware initialization in practice.} In controlled ReLU MLP training (fixed architecture, optimizer, and schedule), Kaiming normal (fan-in) converges faster and with lower loss variance than Xavier normal, empirically corroborating the theory.
(4) \textbf{Transformer variance dynamics.}
In a from-scratch 12-transformer GPT-2--style model during pre-training, I track the layerwise standard deviations of the attention Q/K/V weight matrices.
I observe a pronounced depth-dependent pattern: shallow layers exhibit rapid and sizable early expansion of weight standard deviations, whereas deeper layers expand more gradually and smoothly, with all layers eventually settling into narrow variance bands.
This indicates an emergent depth-dependent variance equilibration: layers closer to the input adapt quickly to high-SNR low-level structure, while deeper layers, constrained by longer residual paths and lower gradient SNR, adjust with smaller effective steps.

\section{Related Work}
Foundational initialization strategies include LeCun’s variance-preserving rule for linear activations \citep{lecun1998gradient}, Glorot/Xavier’s fan-based scaling for tanh/sigmoid \citep{glorot2010understanding}, and He/Kaiming’s rectifier-aware scaling that compensates for the halved second moment under ReLU \citep{he2015delving}. Beyond heuristics, the dynamics near initialization have been formalized via mean-field and signal-propagation analyses—which characterize how activations and gradients evolve through depth and identify “edge-of-chaos” regimes that preserve information flow \citep{schoenholz2017deep,saxe2013exact,yang2020tensor}. In very deep residual stacks, Fixup removes the need for normalization by introducing carefully chosen initialization and residual scaling, enabling stable training even at extreme depths \citep{zhang2019fixup}. For Transformers, a parallel line of work examines how normalization placement and scaling interact with depth: pre-LN improves optimization stability relative to post-LN by smoothing gradient flow \citep{xiong2020layer}, while DeepNet proposes principled depth-scaling rules that allow training on the order of a thousand layers \citep{wang2022deepnet}. At the architectural level, gating choices also affect effective variance and gradient SNR; for example, GLU-style activations have been shown to improve Transformer optimization by modulating signal amplitude \citep{shazeer2020glorot}. From a macro perspective, empirical scaling laws connect model/data/compute to loss, suggesting that stable and data-efficient training at large scale hinges on initialization and normalization choices that keep networks in well-conditioned regimes \citep{kaplan2020scaling,henighan2020scaling}. Finally, practical reports on large language model training underscore the interplay between initialization, precision/quantization, and optimizer settings—e.g., stable pretraining and efficient finetuning with low-precision adapters—further motivating a closer look at initialization in modern LLMs \citep{radford2019language,dettmers2023qlora}.

\section{Theory: Variance in Forward and Backward Passes}

Deep networks train stably when the magnitude (variance) of activations and gradients does not systematically shrink or grow across layers. This section keeps only the essential equations and explains them in plain terms.

\paragraph{Setup.}
Assume each layer has weights with i.i.d.\ zero-mean entries and variance $\sigma_W^2$. Let $x_{l-1}$ be the input to layer $l$, $z_l=W_l x_{l-1}$ the pre-activation, and $x_l=\phi(z_l)$ the post-activation.

\paragraph{Forward pass (keep activations in range).}
For the linear map,
\[
\Var[z_l] = n_{in}\,\sigma_W^2\,\Var[x_{l-1}].
\]
After the nonlinearity $x_l=\phi(z_l)$, the variance is approximately multiplied by a constant that depends on $\phi$ (how much the activation keeps of the signal energy):
\[
\Var[x_l] \approx c_\phi\,n_{in}\,\sigma_W^2\,\Var[x_{l-1}],\qquad c_\phi=\mathbb{E}[\phi(z)^2]/\Var[z].
\]
To avoid vanishing or exploding activations, choose $\sigma_W^2$ such that $\Var[x_l]\approx\Var[x_{l-1}]$, i.e.\ $\sigma_W^2\approx 1/(c_\phi n_{in})$.
Typical cases:
\begin{itemize}
  \item \textbf{ReLU:} roughly half of the mass is active, so $c_\phi\approx 1/2$ and $\sigma_W^2\approx 2/n_{in}$ (He/Kaiming).
  \item \textbf{GELU:} smooth gating gives $c_\phi\approx 0.45\text{--}0.5$, implying a slightly smaller effective gain than ReLU.
\end{itemize}

\paragraph{Backward pass (keep gradients flowing).}
Backprop gives $\delta_{l-1}=W_l^\top(\phi'(z_l)\odot\delta_l)$. Under the same independence approximation,
\[
\Var[\delta_{l-1}] \approx n_{out}\,\sigma_W^2\,d_\phi\,\Var[\delta_l],\qquad d_\phi=\mathbb{E}[\phi'(z)^2].
\]
For ReLU, $\phi'(z)\in\{0,1\}$ with probability $1/2$, so $d_\phi=1/2$. Balancing gradient variance ($\Var[\delta_{l-1}]\approx\Var[\delta_l]$) suggests $\sigma_W^2\approx 2/n_{out}$.

\paragraph{Trade-off and simple choice.}
The forward- and backward-preserving conditions generally cannot be met simultaneously unless $n_{in}\approx n_{out}$ and $c_\phi\approx d_\phi$. In practice with rectifiers it is usually more important to keep the forward signal stable; this explains why fan-in He/Kaiming ($\sigma_W^2\approx 2/n_{in}$) converges faster and more stably than Xavier in my experiments.

\paragraph{Practical notes.}
\begin{itemize}
  \item For ReLU/GELU MLPs, start with fan-in He/Kaiming; if very unbalanced layers cause gradient drift, move slightly toward a fan-average choice.
  \item In deep residual stacks, residual scaling (e.g., $1/\sqrt{L}$) or normalization helps prevent variance drift with depth.
  \item These choices explain the stable standard-deviation band $\sigma\in[10^{-2},10^{-1}]$ observed in E1 and the faster/stabler convergence of He/Kaiming vs.\ Xavier in E2.
\end{itemize}

\paragraph{Implications for Transformers (Q/K/V and MLP).}
Q/K/V projections typically use fan-in with a small nominal std (e.g., $0.02$). If weights grow too large, scaled dot-product attention saturates (few tokens dominate), which implicitly reduces gradient signal-to-noise and pushes scales back toward a stable region. LayerNorm limits forward variance drift by re-centering and re-scaling activations, while AdamW’s preconditioning and weight decay damp gradient variance during backprop. Together these effects yield a depth-dependent equilibrium: shallow layers (high-SNR gradients from inputs) stabilize quickly; deeper layers (longer residual paths, weaker gradients) adjust more gradually—matching the variance bands observed in E3.

\section{Experiments}
This section presents three experiments that progressively connect theory to practice.
\textbf{E1} maps the sensitivity of ReLU MLPs to the initial standard deviation via a logarithmic sweep, identifying vanishing, stable, and exploding regimes.
\textbf{E2} compares Xavier and Kaiming under identical architectures, optimizers, and schedules, testing whether rectifier-aware fan-in scaling improves convergence and stability.
\textbf{E3} trains a 12-layer GPT-2--style model from scratch and tracks layerwise standard deviations for Q/K/V weights, revealing depth-dependent variance bands that emerge during pretraining.

\subsection{E1: Standard-Deviation Sweep (ReLU MLP)}
\textbf{Setup.} I train a ReLU MLP on MNIST for 10 epochs with Adam (learning rate $0.01$). The network is a $784\rightarrow64\rightarrow32\rightarrow32\rightarrow10$ classifier. I sweep $25$ initial weight standard deviations,
with $\sigma$ taking 25 values \emph{logarithmically spaced} between $10^{-4}$ and $10$, reinitializing the model for each $\sigma$. I record epoch loss and classification accuracy.

\textbf{Protocol.} For every $\sigma$ in the sweep: initialize all linear layers with the specified std, train for a fixed schedule, and log the final accuracy and loss trajectory. This isolates the effect of the global scale while holding data, model, optimizer, and schedule fixed.

\textbf{Findings.} Extremely small scales ($\sigma \lesssim 10^{-3}$) yield vanishing updates and poor accuracy, consistent with attenuated forward/gradient variance. Very large scales ($\sigma \gtrsim 1$) produce unstable loss and occasional divergence, indicating exploding activations/gradients. A broad \emph{stable band} emerges for
$\sigma \in [10^{-2}, 10^{-1}]$: training is smooth, gradients remain well-behaved, and accuracy peaks within this interval. These observations match the variance-preservation view: scales that roughly keep layer-to-layer variance constant avoid both saturation and blow-up and lead to faster effective learning.

\begin{figure}[t]
\centering
\includegraphics[width=0.95\columnwidth]{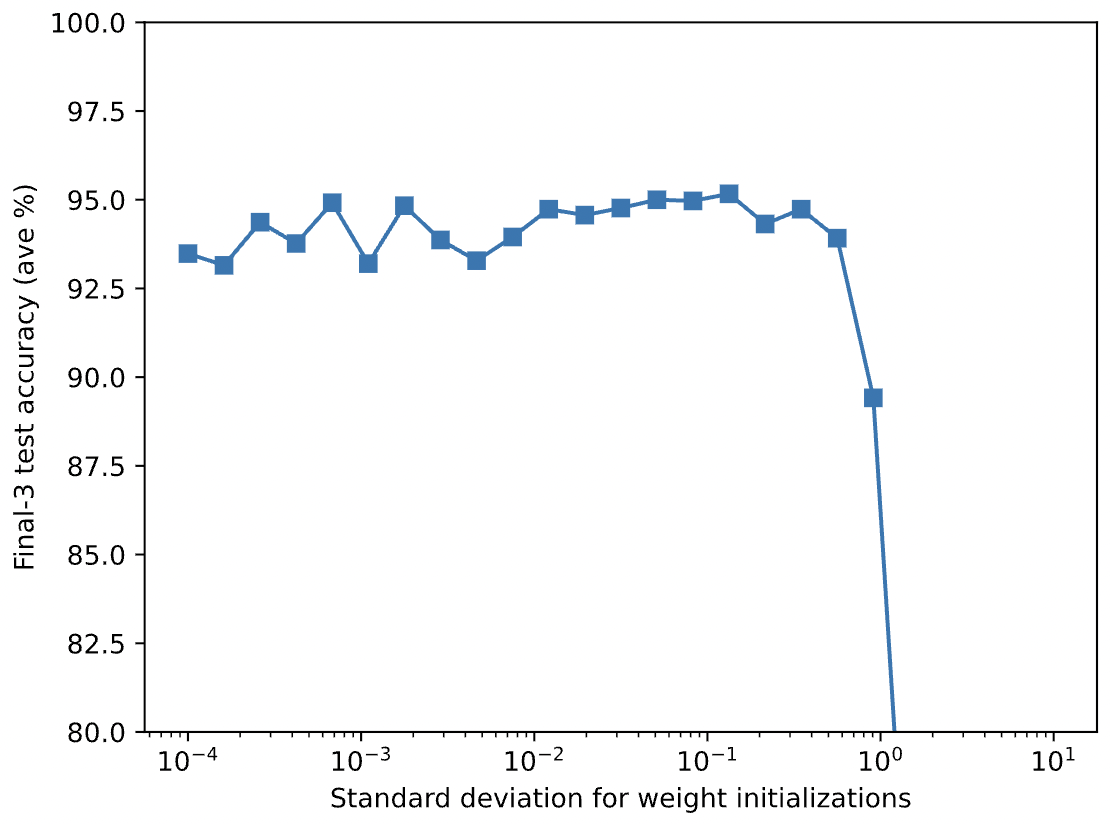}
\caption{E1: Loss trajectories for selected initialization scales. Stable learning occurs for $\sigma \in [10^{-2}, 10^{-1}]$, while too small or too large $\sigma$ causes vanishing or unstable behavior.}
\end{figure}

\subsection{E2: Xavier vs.\ Kaiming under ReLU}

\textbf{Setup.} I train a model for a binary classification task on the UCI Wine dataset. The network is $11\rightarrow16\rightarrow32\rightarrow32\rightarrow1$ with ReLU after each hidden layer. I compare two initialization schemes under identical data, architecture, optimizer, and schedule: \emph{Xavier normal} versus \emph{Kaiming (fan-in, nonlinearity=\texttt{relu})}, implemented as PyTorch's \texttt{kaiming\_uniform\_} for linear layers. To assess robustness, I repeat training across $10$ randomized runs and log training loss, validation accuracy, and simple stability indicators (loss spikes, gradient norm excursions).

\textbf{Protocol.} For each run and each initialization: (1) reinitialize all linear layers with the scheme under test; (2) train for the fixed schedule; (3) record the full loss trajectory, time-to-target metrics (e.g., epochs to reduce loss by $95\%$ from its initial value), final validation accuracy, and the standard deviation of the per-epoch loss (as a measure of optimization smoothness). This design isolates the effect of the initialization while controlling for every other factor.

\textbf{Findings.} Kaiming consistently outperforms Xavier:
(i) \emph{Faster convergence}—the median epochs-to-target is lower with Kaiming, and the early loss slope is steeper;
(ii) \emph{Comparable or better accuracy}—final validation accuracy matches or slightly exceeds Xavier on average.
Taken together, these results align with the variance-propagation analysis: rectifier-aware fan-in scaling preserves forward signal magnitude more faithfully, easing gradient flow and yielding quicker, more stable optimization under ReLU.

\begin{figure}[t]
  \centering
  \includegraphics[width=0.95\columnwidth]{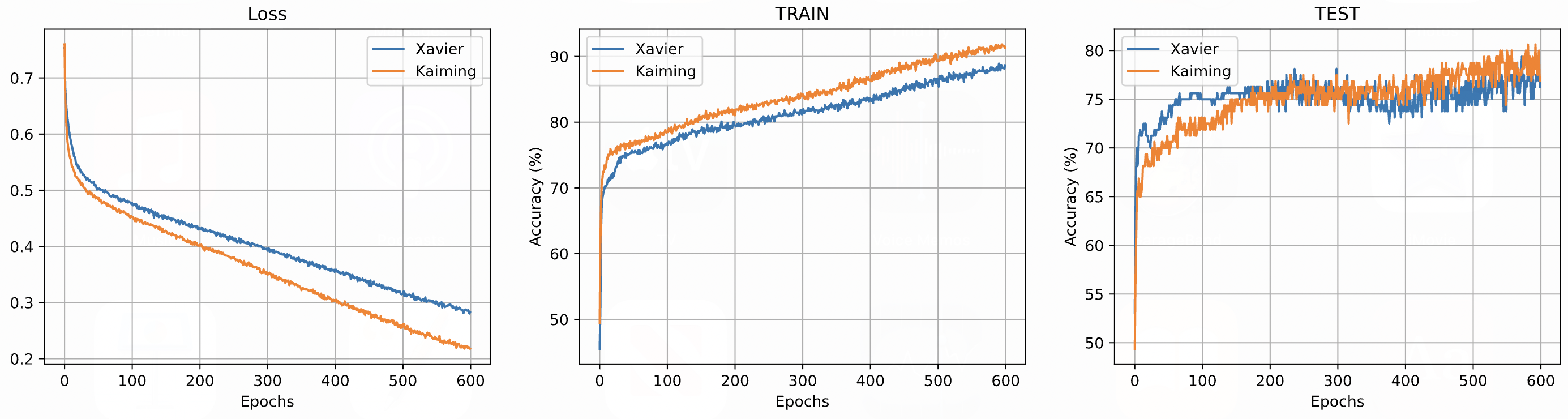}
  \caption{E2: Loss, train accuracy, and test accuracy for Xavier (blue) vs.\ Kaiming (orange) under identical settings. Kaiming shows faster decay and higher accuracy on the training dataset.}
\end{figure}

\begin{figure}[t]
  \centering
  \includegraphics[width=0.95\columnwidth]{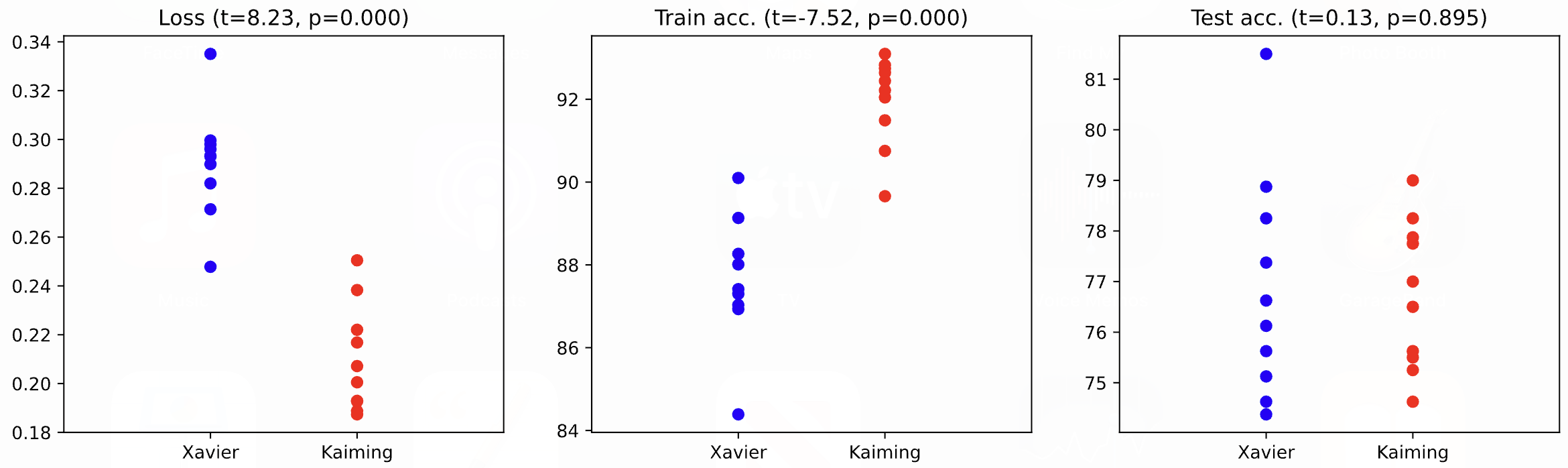}
  \caption{E2 (t-test): Aggregated statistics over 10 runs.
  Paired t-tests show significant differences between Xavier and Kaiming
  in both loss and training accuracy ($p < 0.05$).}
\end{figure}

\subsection{E3: Variance Dynamics in a GPT-2 Model}

\textbf{Setup.}
To examine how variance evolves in deep transformers, I pretrain a 12-transformer GPT-2--style model built from scratch. Each transformer block includes multi-head self-attention and a feed-forward MLP with GELU activation. The model is initialized with a standard configuration (\texttt{normal} distribution, std $0.02$ for most modules and \texttt{xavier\_normal} for the embedding matrix) and trained with AdamW at learning rate $1\times10^{-4}$, batch size $16$. The experiment tracks the standard deviation of weights in all layers of attention Q/K/V projections.

\textbf{Protocol.}
At each training checkpoint, I (i) compute the standard deviation of the weights for every layer and module (attention Q/K/V) to obtain a depth-wise time series of layerwise stds; and (ii) fix a representative layer (e.g., a specific attention projection) and log its empirical weight distribution (histogram) across training steps.
This instrumentation simultaneously captures depth-dependent variance adaptation (via per-layer std trajectories) and within-layer distributional evolution over time.

\begin{figure}[t]
  \centering
  \includegraphics[width=0.95\columnwidth]{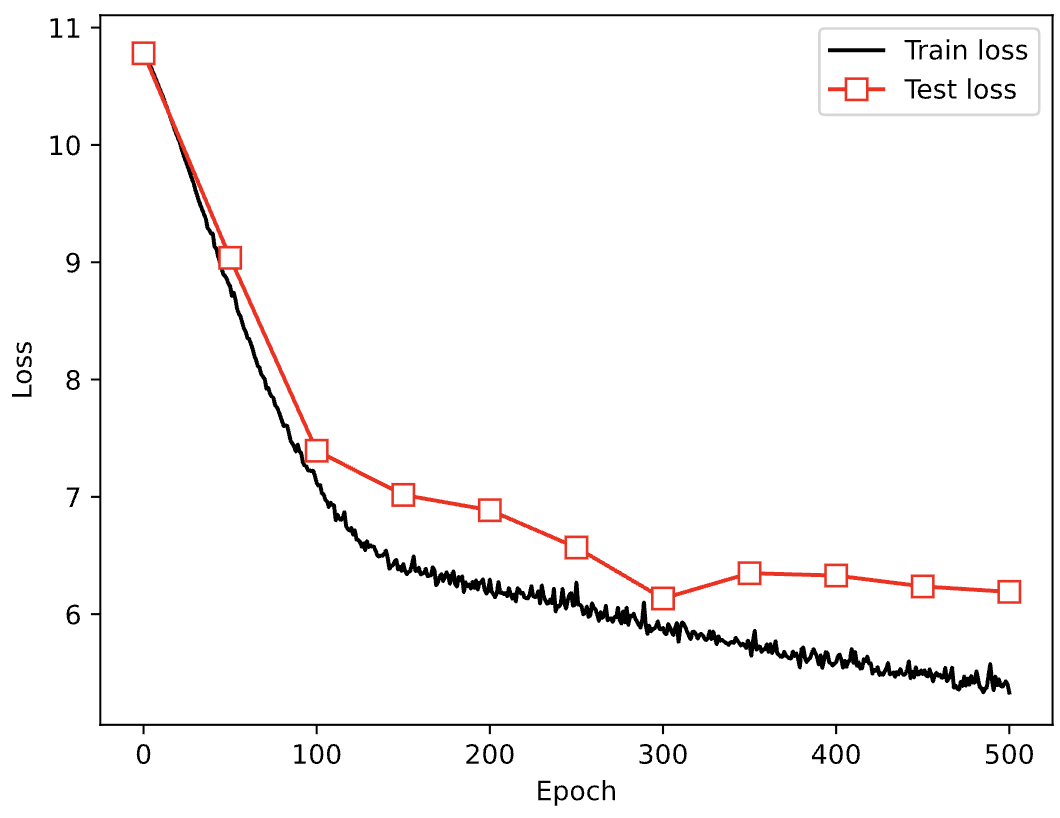}
\caption{E3: Loss curves (train vs.\ test) during pre-training of a from-scratch GPT-2 model.}
\end{figure}

\begin{figure}[t]
  \centering
  \includegraphics[width=0.95\columnwidth]{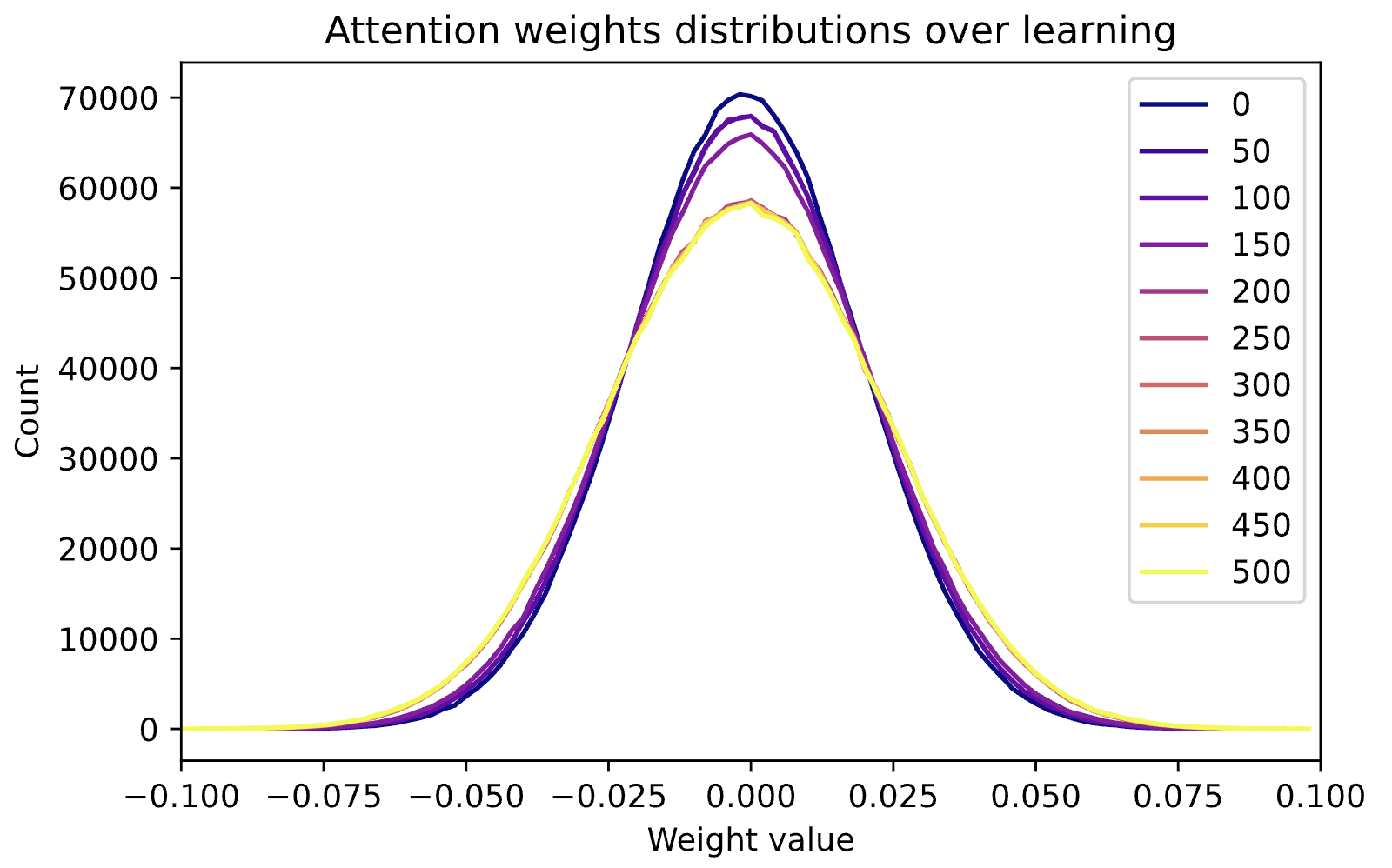}
\caption{E3 (one sample layer's weight distribution): Weight-distribution dynamics sampled every 50 epochs. The distributions become increasingly sparse over time, with mass progressively concentrating near zero.}
\end{figure}

\begin{figure}[t]
  \centering
  \includegraphics[width=0.95\columnwidth]{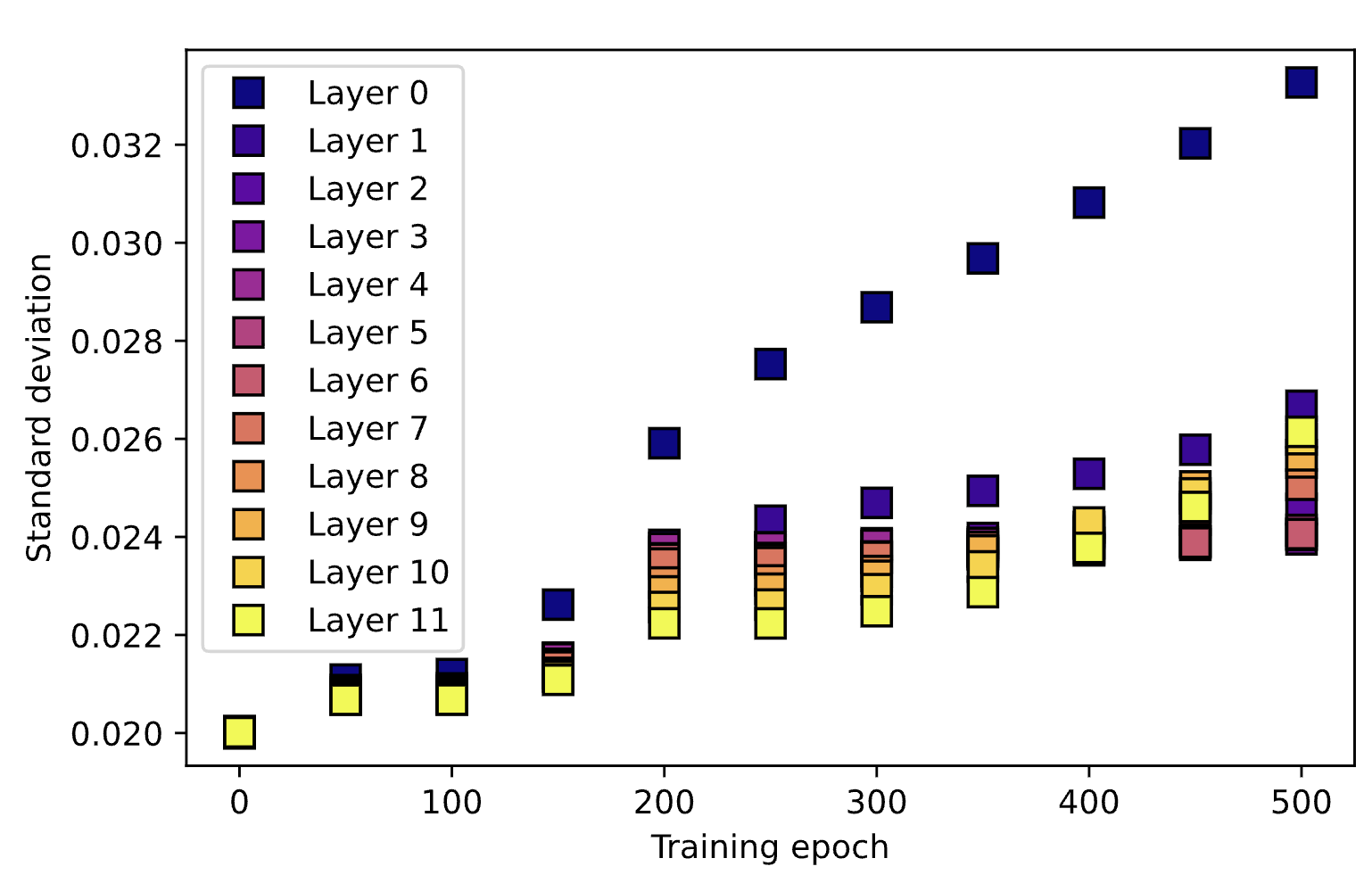}
\caption{E3: Tracking layerwise Q/K/V standard deviations across all 12 transformer blocks during pre-training. Each layer exhibits distinct variance trajectories, reflecting depth-dependent adaptation and equilibration.}
\end{figure}

\textbf{Findings.}
After pretraining, the model exhibits clear depth-dependent variance dynamics.
For attention Q/K/V, the lower layers show rapid and pronounced expansion of weight standard deviation during early training—std values increase sharply within the first few thousand iterations, reflecting strong adaptation to raw input statistics.
In contrast, higher layers expand much more slowly, with smaller and smoother std growth, indicating that their weights adjust gradually under weaker, residual-gradient signals constrained by normalization.
When visualizing individual weight matrices, the post-training distributions become noticeably sparser: many near-zero entries remain, while a small fraction of large-magnitude weights dominate, suggesting selective amplification of salient connections.
Overall, the variance trajectories reveal a progressive equilibration pattern—fast and volatile adaptation in shallow layers, slower convergence in deeper layers, and eventual stabilization of each layer’s variance around a narrow equilibrium band.
This asymmetric evolution highlights how transformer depth shapes the flow of statistical adaptation during pretraining.

\section{Conclusion and Outlook}
This paper connects classical variance-propagation analysis with empirical behavior in modern transformers.
Experiment~E1 maps the effect of the initial standard deviation and identifies a broad but sensitive stability band $\sigma\in[10^{-2},10^{-1}]$ for ReLU MLPs.
Experiment~E2 shows that under controlled conditions Kaiming (fan-in, rectifier-aware) initialization converges faster and with lower loss variability than Xavier, corroborating the forward--backward variance trade-off discussed in the theory section.
Experiment~E3 extends these insights to a GPT-2--style model, revealing depth-dependent variance dynamics and distributional changes in attention and MLP weights.

A prominent observation in E3 is the asymmetric evolution of Q/K/V parameters across depth: lower layers exhibit rapid and pronounced growth in weight standard deviation early in training, while higher layers expand more slowly and with smaller amplitude.
I attribute this pattern to the interaction of (i) higher gradient signal-to-noise near the input, which encourages aggressive early scaling in shallow blocks; (ii) residual path length and pre-normalization, which temper effective step sizes in deeper blocks; and (iii) attention softmax saturation and weight decay in AdamW, which impose implicit constraints that discourage large parameter scales as depth increases.
Post-training, weight distributions become sparser—many coefficients remain near zero while a minority grow in magnitude—indicating selective amplification of salient connections.
Together, these findings support the view that transformers self-organize into layer-specific, depth-dependent variance regimes: fast, volatile adaptation in shallow layers and slower, damped equilibration in deeper layers.

\paragraph{Practical implications.}
The results motivate simple, actionable guidance: prefer fan-in Kaiming for rectifier networks; initialize transformer projections with small std (e.g., 0.02) and monitor per-layer standard deviations and gradient norms; if shallow layers fail to expand or deep layers stagnate, modest adjustments to warmup, residual scaling, or weight decay typically restore healthy variance flow.
These diagnostics complement the stability band from E1 and the rectifier-aware advantage from E2.

\paragraph{Outlook.}
Several extensions follow naturally:
\begin{itemize}
  \item \textbf{Adaptive, depth-aware initialization.} Learn per-layer (or per-head) scales so that shallow layers start closer to their eventual variance levels while deeper layers start more conservatively, reducing early transients.
  \item \textbf{Optimizer and schedule coupling.} Jointly tune warmup length, weight decay, and gradient clipping with initialization to maintain stable variance under different batch sizes and sequence lengths.
  \item \textbf{Scaling with depth and width.} Evaluate whether the depth-dependent equilibration persists under larger models and datasets, and compare to alternative parameterizations (e.g., $\mu$P) and depth-scaling rules.
\end{itemize}

\bibliographystyle{plainnat}
\bibliography{references}

\begin{thebibliography}{14}
\providecommand{\natexlab}[1]{#1}
\providecommand{\url}[1]{\texttt{#1}}
\expandafter\ifx\csname urlstyle\endcsname\relax
  \providecommand{\doi}[1]{doi: #1}\else
  \providecommand{\doi}{doi: \begingroup \urlstyle{rm}\Url}\fi

\bibitem[Dettmers et~al.(2023)Dettmers, Lewis, Shleifer, and Zettlemoyer]{dettmers2023qlora}
Tim Dettmers, Mike Lewis, Sam Shleifer, and Luke Zettlemoyer.
\newblock Qlora: Efficient finetuning of quantized large language models.
\newblock \emph{arXiv preprint arXiv:2305.14314}, 2023.

\bibitem[Glorot and Bengio(2010)]{glorot2010understanding}
Xavier Glorot and Yoshua Bengio.
\newblock Understanding the difficulty of training deep feedforward neural networks.
\newblock In \emph{Proceedings of the Thirteenth International Conference on Artificial Intelligence and Statistics (AISTATS)}, pages 249--256, 2010.

\bibitem[He et~al.(2015)He, Zhang, Ren, and Sun]{he2015delving}
Kaiming He, Xiangyu Zhang, Shaoqing Ren, and Jian Sun.
\newblock Delving deep into rectifiers: Surpassing human-level performance on imagenet classification.
\newblock In \emph{Proceedings of the IEEE International Conference on Computer Vision (ICCV)}, pages 1026--1034, 2015.

\bibitem[Henighan et~al.(2020)Henighan, Kaplan, Katz, Chen, et~al.]{henighan2020scaling}
Tom Henighan, Jared Kaplan, Mayur Katz, Mark Chen, et~al.
\newblock Scaling laws for autoregressive generative modeling.
\newblock \emph{arXiv preprint arXiv:2010.14701}, 2020.

\bibitem[Kaplan et~al.(2020)Kaplan, McCandlish, Henighan, Brown, et~al.]{kaplan2020scaling}
Jared Kaplan, Sam McCandlish, Tom Henighan, Tom~B. Brown, et~al.
\newblock Scaling laws for neural language models.
\newblock \emph{arXiv preprint arXiv:2001.08361}, 2020.

\bibitem[LeCun et~al.(1998)LeCun, Bottou, Orr, and M{\"u}ller]{lecun1998gradient}
Yann LeCun, L{\'e}on Bottou, Genevieve~B. Orr, and Klaus-Robert M{\"u}ller.
\newblock Efficient backprop.
\newblock In Genevieve~B. Orr and Klaus-Robert M{\"u}ller, editors, \emph{Neural Networks: Tricks of the Trade}. Springer, 1998.

\bibitem[Radford et~al.(2019)Radford, Wu, Child, Luan, Amodei, and Sutskever]{radford2019language}
Alec Radford, Jeffrey Wu, Rewon Child, David Luan, Dario Amodei, and Ilya Sutskever.
\newblock Language models are unsupervised multitask learners.
\newblock OpenAI Blog, 2019.
\newblock Technical report.

\bibitem[Saxe et~al.(2013)Saxe, McClelland, and Ganguli]{saxe2013exact}
Andrew~M. Saxe, James~L. McClelland, and Surya Ganguli.
\newblock Exact solutions to the nonlinear dynamics of learning in deep linear neural networks.
\newblock \emph{arXiv preprint arXiv:1312.6120}, 2013.

\bibitem[Schoenholz et~al.(2017)Schoenholz, Gilmer, Ganguli, and Sohl-Dickstein]{schoenholz2017deep}
Samuel~S. Schoenholz, Justin Gilmer, Surya Ganguli, and Jascha Sohl-Dickstein.
\newblock Deep information propagation.
\newblock \emph{arXiv preprint arXiv:1611.01232}, 2017.

\bibitem[Shazeer(2020)]{shazeer2020glorot}
Noam Shazeer.
\newblock Glu variants improve transformer.
\newblock \emph{arXiv preprint arXiv:2002.05202}, 2020.

\bibitem[Wang et~al.(2022)]{wang2022deepnet}
Shuming Wang et~al.
\newblock Deepnet: Scaling transformers to 1,000 layers.
\newblock \emph{arXiv preprint arXiv:2203.00555}, 2022.

\bibitem[Xiong et~al.(2020)Xiong, Yang, He, Zheng, Zheng, Xing, Zhang, and Sun]{xiong2020layer}
Rui Xiong, Yunchang Yang, Shijie He, Kai Zheng, Qiao Zheng, Chao Xing, Tong Zhang, and Zhihua Sun.
\newblock On layer normalization in the transformer architecture.
\newblock \emph{arXiv preprint arXiv:2002.04745}, 2020.

\bibitem[Yang(2020)]{yang2020tensor}
Greg Yang.
\newblock Tensor programs i: Wide neural networks are gaussian processes.
\newblock \emph{arXiv preprint arXiv:1910.12478}, 2020.

\bibitem[Zhang et~al.(2019)Zhang, Dauphin, and Ma]{zhang2019fixup}
Hongyi Zhang, Yann~N. Dauphin, and Tengyu Ma.
\newblock Fixup initialization: Residual learning without normalization.
\newblock In \emph{International Conference on Learning Representations (ICLR)}, 2019.

\end{thebibliography}
\end{document}